\documentclass[letterpaper]{article} 
\usepackage{aaai25}  
\usepackage{times}  
\usepackage{helvet}  
\usepackage{courier}  
\usepackage[hyphens]{url}  
\usepackage{graphicx} 
\usepackage{natbib}  
\usepackage{caption} 
\usepackage{algorithm}
\usepackage{algorithmic}
\usepackage{pifont}
\usepackage{multirow}
\usepackage{booktabs}
\usepackage{newfloat}
\usepackage{listings}
\DeclareCaptionStyle{ruled}{labelfont=normalfont,labelsep=colon,strut=off} 
\floatstyle{ruled}
\newfloat{listing}{tb}{lst}{}
\floatname{listing}{Listing}
\usepackage[table]{xcolor}

\title{\includegraphics[height=1.5\baselineskip]{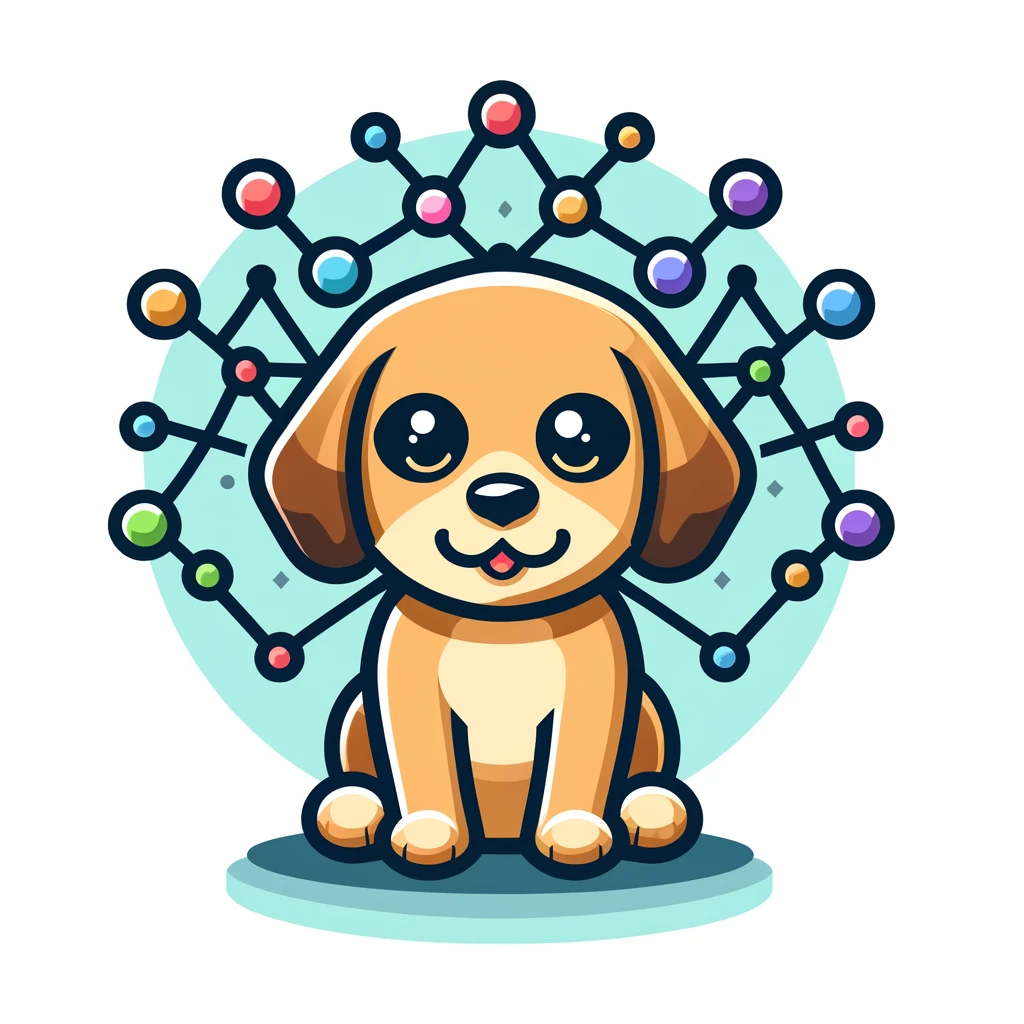}Debate on Graph: \\ a Flexible and Reliable Reasoning Framework for Large Language Models}
\author{
    Jie Ma\textsuperscript{\rm 1}\thanks{Corresponding author.},
    Zhitao Gao\textsuperscript{\rm 1},
    Qi Chai\textsuperscript{\rm 2},
    Wangchun Sun\textsuperscript{\rm 1},
    Pinghui Wang\textsuperscript{\rm 1},
    Hongbin Pei\textsuperscript{\rm 1},
    Jing Tao\textsuperscript{\rm 1},
    Lingyun Song\textsuperscript{\rm 3},
    Jun Liu\textsuperscript{\rm 1},
    Chen Zhang\textsuperscript{\rm 4},
    Lizhen Cui\textsuperscript{\rm 5}
}
\affiliations{
    \textsuperscript{\rm 1} MOE KLINNS Lab, Xi'an Jiaotong University \\
    \textsuperscript{\rm 2} The Hong Kong University of Science and Technology (Guangzhou)\\
    \textsuperscript{\rm 3} Northwestern Polytechnical University\\
    \textsuperscript{\rm 4} Zhejiang Createlink Technology\\
    \textsuperscript{\rm 5} Shandong University\\

    jiema@xjtu.edu.cn
}

\definecolor{darkgreen}{rgb}{0.0, 0.5, 0.0}
\newcommand{\ts}[1]{\textsuperscript{\textcolor{darkgreen}{#1}}}

\begin{document}

\maketitle

\begin{abstract}
Large Language Models (LLMs) may suffer from hallucinations in real-world applications due to the lack of relevant knowledge. In contrast, knowledge graphs encompass extensive, multi-relational structures that store a vast array of symbolic facts. Consequently, integrating LLMs with knowledge graphs has been extensively explored, with Knowledge Graph Question Answering (KGQA) serving as a critical touchstone for the integration. This task requires LLMs to answer natural language questions by retrieving relevant triples from knowledge graphs. However, existing methods face two significant challenges: \textit{excessively long reasoning paths distracting from the answer generation}, and \textit{false-positive relations hindering the path refinement}. In this paper, we propose an iterative interactive KGQA framework that leverages the interactive learning capabilities of LLMs to perform reasoning and Debating over Graphs (DoG). Specifically, DoG employs a subgraph-focusing mechanism, allowing LLMs to perform answer trying after each reasoning step, thereby mitigating the impact of lengthy reasoning paths. On the other hand, DoG utilizes a multi-role debate team to gradually simplify complex questions, reducing the influence of false-positive relations. This debate mechanism ensures the reliability of the reasoning process. Experimental results on five public datasets demonstrate the effectiveness and superiority of our architecture. Notably, DoG outperforms the state-of-the-art method ToG by 23.7\% and 9.1\% in accuracy on WebQuestions and GrailQA, respectively. Furthermore, the integration experiments with various LLMs on the mentioned datasets highlight the flexibility of DoG. Code is available at \url{https://github.com/reml-group/DoG}.
\end{abstract}

\section{Introduction}
Large Language Models (LLMs), characterized by their substantial parameter amount \cite{zhang2023siren} and training on extensive, diverse, and unlabeled data \cite{rawte2023survey}, exhibit remarkable proficiency in a wide range of natural language understanding and generation tasks \cite{lin2023techs, liu2024survey}. For example, GPT-4 \cite{achiam2023gpt} demonstrates human-level performance across a majority of professional and academic exams originally intended for humans. However, recent studies \cite{guan2024mitigating, waldendorf2024, gunjal2024detecting} have revealed that they may suffer from hallucinations in real-world applications due to a deficiency in relevant knowledge.

Knowledge graphs \cite{wang2024knowledge} are large-scale, multi-relational structures housing a plethora of symbolic facts, such as the triple \texttt{\textless The Eiffel Tower, locatedIn, Paris\textgreater}. The incorporation of these structured facts may tackle the aforementioned issue of hallucinations in LLMs \cite{guan2024mitigating, quintero2024, shi2023hallucination}. One approach to evaluating the integration of knowledge graphs with LLMs is through Knowledge Graph Question Answering (KGQA) \cite{9416312}, which requires machines to answer natural language questions by retrieving relevant facts from knowledge graphs. Recent works \cite{li2024flexkbqa, toroghi2024right, nie2024code} primarily follow an iterative inference paradigm, consisting of two steps: (1) identifying the initial entity in the question, and (2) retrieving and refining the inference path iteratively until reaching the answer or obtaining sufficient evidence to answer the question. Although they have achieved significant success, they still suffer from \textit{excessively long paths} and \textit{false-positive relations}.
\begin{figure}[tbp]
	\centering  
	\includegraphics[width=\linewidth]{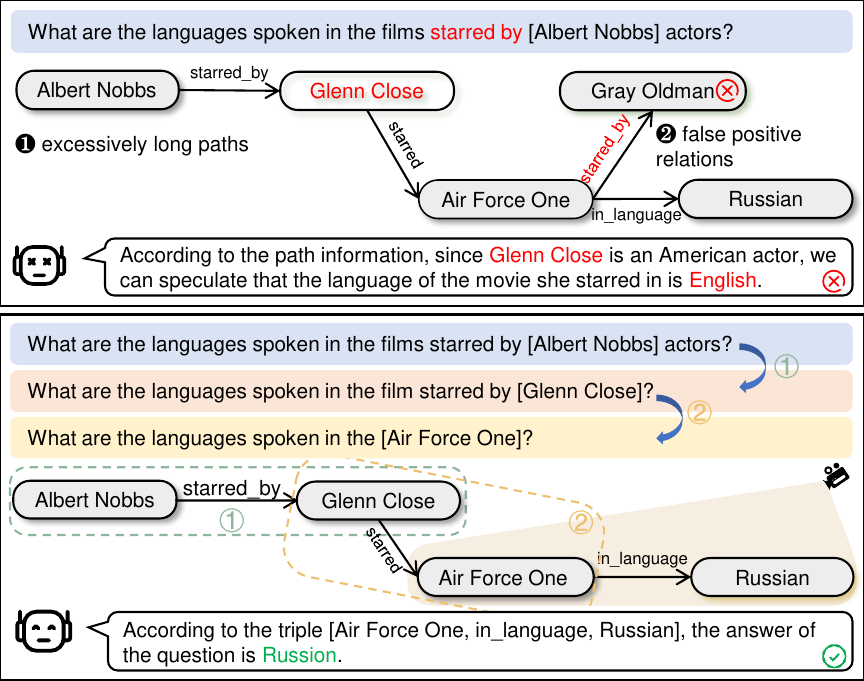}
	\caption{Illustration of challenges and our solutions.}
	\label{fig:comp-case}
\end{figure}

\textbf{Challenge 1:} excessively long paths distracting from the answer generation. Existing methods \cite{ye2022rng, guo2023, kim2023kg} usually feed a lengthy evidence path like \{\texttt{\textless Albert Nobbs, starred\_by, Glenn Close\textgreater}, $\cdots$,  \texttt{\textless Air Force One, starred\_by, Gray Oldman\textgreater}, $\cdots$\} at the top of Fig. \ref{fig:comp-case} into LLMs to perform answer generation in a single step, which may make it challenging for LLMs to discern the key points in the path. For instance, LLMs may focus on the tail entity \texttt{Glenn Close} and employ their internal prior knowledge to generate answers. This will result in answers that appear reasonable but are incorrect.

\textbf{Challenge 2:} false-positive relations hindering the path refinement. Current methods \cite{bai2023complex, hubench, li2024flexkbqa} typically focus on identifying relations within graphs that closely match or have the same meaning as those in the questions, even if the relations have already been identified in previous reasoning steps. For example, at the top of Fig. \ref{fig:comp-case}, these methods may select \texttt{starred\_by}, which was used in the previous reasoning step and is mentioned in the question, to expand paths rather than choosing \texttt{in\_language} when dealing with the entity \texttt{Air Force One}. This will lead to incomplete evidence paths.

To address these challenges, we propose an iterative interactive KGQA framework that leverages the interactive learning capabilities of LLMs to perform reasoning and \textbf{D}ebating \textbf{o}ver \textbf{G}raphs, dubbed \textbf{DoG}. Unlike existing approaches \cite{jiang2023structgpt, luo2023reasoning, sun2024think} that seek to construct a complete evidence chain before answering questions, our architecture employs a subgraph-focusing mechanism that allows LLMs to perform answer trying after each reasoning step. For each filtered triple, DoG uses LLMs to assess whether sufficient information is available to answer the current question. In this way, the triple in each reasoning step, such as \texttt{\textless Glenn Close, starred, AirForce One\textgreater} in the bottom of Fig. \ref{fig:comp-case}, can be deeply pondered by LLMs. If the triple does not support answering the current question, DoG employs a multi-role LLM team to debate and simplify the question based on the triple. The iterative process allows complex multi-hop questions to be gradually transformed into single-hop questions, which enables LLMs not to be disturbed by the relation that is retrieved in the previous reasoning step. For example, the relation \texttt{starred\_by} that is linked with \texttt{Air Force One} will not disturb reasoning after the simplification procedure \ding{173}. This is inspired by the human brain in tackling complex questions, which guides LLMs to reason on graphs through chain-of-thought \cite{wei2022chain}. The simplification process can also enhance the transparency of the reasoning process. 

To verify the effectiveness and superiority of our architecture, we conduct thorough experiments on five public KGQA datasets: MetaQA \cite{zhang2018variational}, WebQSP \cite{yih2016value}, CWQ \cite{talmor2018web}, WebQuestions \cite{berant2013semantic}, and GrailQA \cite{gu2021beyond}. Our findings show that DoG achieves state-of-the-art results on all datasets, except for the 2-hop and 3-hop questions within MetaQA. Notably, DoG outperforms the strong baseline ToG \cite{sun2024think} by 23.7\% and 9.1\% in accuracy on WebQuestions and GrailQA, respectively. In summary, our contributions are threefold.
\begin{itemize}
    \item We propose a flexible and reliable reasoning framework, DoG, which enables LLMs to reason and debate over knowledge graphs and answer questions after thorough deliberation.
    \item We introduce a strategy, which transforms questions from complex to easy through the interactive learning of a multi-role LLM team, for handling complex reasoning on knowledge graphs. This guides LLMs to engage in step-by-step reasoning, thereby enhancing the reliability of the reasoning process.
    \item Extensive experiments and ablation studies are carried out on five public datasets to demonstrate the effectiveness and superiority of our architecture. Furthermore, we also conduct integration experiments with various LLMs to verify the flexibility of DoG.
\end{itemize}

\section{Related Work}

The methods of LLM reasoning over knowledge graphs can be classified into \emph{batch triple recalling}, and \emph{reasoning path refining} from the perspective of evidence gathering.

\textbf{Batch triple recalling.} Knowledge graphs typically store an extensive amount of facts \cite{cui2023lifelong}. For instance, Freebase \cite{Bollacker2008FreebaseAC} contains over 1.9 billion triples, and even the smaller non-open-domain MetaQA \cite{zhang2018variational} includes over 130,000 triples. The number of relevant triples can be substantial even when constrained by the entities in a given question. Injecting all these triples into the context window of LLMs to perform reasoning not only incurs a high encoding cost but also introduces significant noise \cite{wei2023kicgpt}. To address this issue, previous studies \cite{shu-etal-2022-tiara, ye2022rng,guo2023} focus on how to filter suitable facts. For instance, KAPING \cite{baek2023knowledge} projects questions and triples into the same space to obtain relevant knowledge by semantic similarity. KG-GPT \cite{kim2023kg} further focuses on fine-grained question representations, decomposing multi-hop questions into sub-questions and matching the relations associated with entities in those sub-questions, then selecting the top-k relevant relations to form evidence triples. Similarly, KGR \cite{guan2024mitigating} splits the retrieved triples into several chunks and utilizes LLM to distinguish the critical triple relevant with questions.

\textbf{Reasoning path refining.} The paradigm of this kind of method \cite{gu2023don,jiang2023structgpt, liu2023agentbench, luo2023reasoning, sun2024think,guo2023knowledgenavigator} is first to identify the initial entity in the question, then to iteratively retrieve and refine the reasoning path until reaching the answer or obtaining sufficient evidence to answer the question, and finally to employ LLMs to generate answers based on the refined path. For example, \citet{jiang2023structgpt} proposed an iterative reading-reasoning approach, which iterates an invoking-linearization-generation procedure. It utilizes LLMs to perform reasoning on the interface that is specifically designed for reading structured data until deriving the final answer. Similarly, \citet{sun2024think} introduced a deep and responsible reasoning framework, which first conducts a beam search on a graph from the entity within questions and then acquires multiple reasoning paths as evidence for answer generation. It is noteworthy that these methods all treat the LLM as a tool for accomplishing specific tasks, conceptualizing it as function executors, and relying on in-context learning \cite{dong2022survey} or fine-tuning to refine its outputs \cite{jiang2024kg}. However, some studies \cite{zhao2024expel,zhang2023exploring,schumann2024velma} have demonstrated that LLMs can be induced to exhibit human personality traits and role distinctions to undertake complex reasoning tasks.

\textbf{Communicative Agents.} The primary objective of agents is to collaboratively address complex tasks in a productive and efficient manner through autonomous communication and negotiation \cite{chan2023chateval, liang2023encouraging, yang2023how2comm, kirk2024improving}. LLMs such as ChatGPT and Vicuna \cite{vicuna2023} are frequently employed as these communicative agents. Recently, numerous studies have investigated the application of these agents in various domains, including AI societies \cite{LiHIKG23}, software development \cite{qian2023communicative}, translation \cite{liang2023encouraging}, arithmetic problem-solving \cite{du2023improving}, dialogue response generation \cite{chan2023chateval}, and strategic planning among robots \cite{mandi2023roco}. Specifically, \citet{wang2023can} guided ChatGPT to emulate expert system reviewers, thereby improving the quality of its literature retrieval queries. \citet{kong2023better} introduced a strategically designed role-playing prompt method to enhance reasoning abilities by assigning appropriate expert roles for tasks. Additionally, \citet{shen2024decision} assessed the changes in decision-making abilities when LLM assumes different personality traits. Inspired by these studies, we explore the benefit of multi-agent role differentiation and debates for complex reasoning on knowledge graphs.

\section{Method}

\subsection{Task Formulation}
Given a knowledge graph $\mathcal{G}$ consisting of $N$ triples, represented as $\{ (e_i^{l}, r_l, e_{i+1}^{l}) | e_i \in \mathcal{E}, r_l \in \mathcal{R}, i \in [1, I], l \in [1, L] \}$, where $e_i^{l}$ and $e_{i+1}^{l}$ denote the head and tail entity, respectively, $I$ is the number of entities, $L$ denotes the number of relations, and $r_l$ is the relation between entities, KGQA requires machines to answer natural language questions $q$ based on retrieved evidence paths $P = \{p_j\}_{j=1}^m$ with $p_j$ representing a triple and $m$ denoting the number of triples. In this paper, we leverage LLMs to reason over $P$ and generate answers $\hat{a}$ word by word.

\subsection{Overview}
\begin{figure*}[htbp]
\centering  
\includegraphics[width=\linewidth]{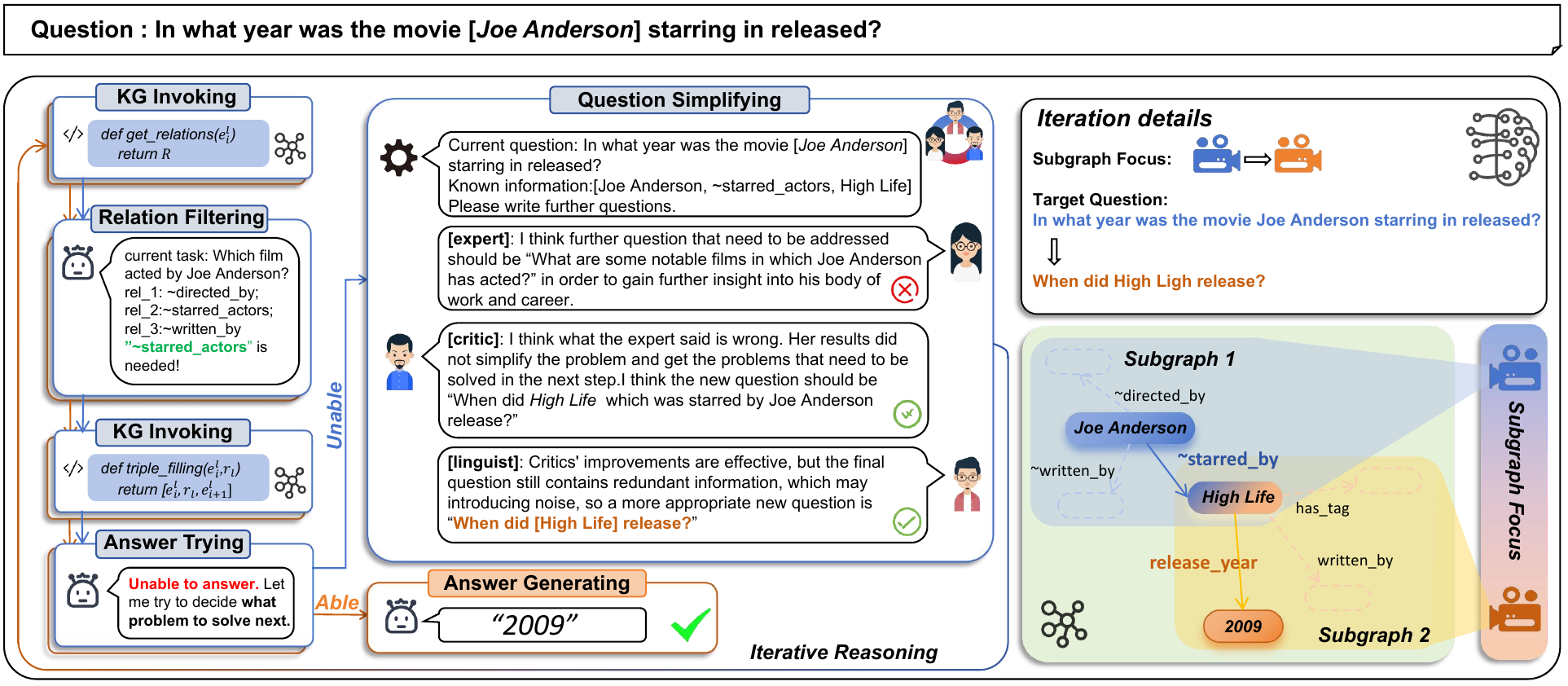}
\caption{DoG framework. Given a question, our framework first enables LLMs to interact with knowledge graphs to retrieve the most relevant triple. Subsequently, it employs a subgraph-focusing mechanism, allowing LLMs to attempt answering at each reasoning step. If further reasoning is required, DoG leverages a multi-role LLM team to simplify the question from complex to easy based on the retrieved triples.}
\label{fig:arc}
\end{figure*}

As depicted in Fig. \ref{fig:arc}, given a $K$-hop question $q$ and the initial topic entity $e_i^{l}$ within $q$, our framework first invokes knowledge graphs to retrieve the set of candidate relations $R$ linked to $e_i^{l}$. Then, it enables LLMs to filter out the most relevant relation $\hat{r}_l$ from $R$ based on in-context learning. Subsequently, the knowledge graph is invoked again to complete the triple information from $(e_i^{l}, \hat{r}_l, ?)$ to $(e_i^l, \hat{r}_l, e_{i+1}^l)$. Fourthly, DoG focuses on the current reasoning state and employs LLMs to decide on the subsequent action based on the completed triple: providing a direct answer to the question or performing deep thinking with further iterations. In the latter scenario, a multi-role LLM team leverages the mentioned triple to transform the $K$-hop question to a $K$-1 hop (slightly easier) one through debate, with the tail entity $e_{i+1}^l$ being the subsequent topic entity for the simplified question in the next iteration. All of these debate steps are autonomously executed by the LLM team. The iteration will be ended until LLMs generate answers in the fourth step.

\subsection{Knowledge Graph Invoking}
Reasoning on graphs requires LLMs first to identify relevant knowledge triples. To facilitate this, we have designed two interactive interfaces specifically tailored to retrieve these triples from knowledge graphs. The interfaces are invoked as needed, depending on the requirements.
\begin{itemize}
    \item \textit{get\_relations}($e_i^{l}$): This interface is designed to retrieve the candidate relation set $R$ associated with the entity $e_i^{l}$. For example, in Fig. \ref{fig:arc}, it is invoked to retrieve the candidate relation set of \texttt{Joe Anderson}.
    \item \textit{triple\_filling}($e_i^{l}$, $\hat{r}_l$): This interface is responsible for obtaining the tail entity \textless$e_i^{l}, \hat{r}_l, ?$\textgreater~given the head entity and the filtered relation. We will introduce relation filtering in the next subsection.
\end{itemize}
The underlying mechanisms of these interfaces are implemented through either SPARQL (for Freebase queries) or specific matching (for questions in MetaQA). To facilitate comprehension and generation by LLMs, all entities and relationships above the interfaces are expressed in natural language, with the conversion between a Machine ID (MID) and a corresponding friendly name carried out exclusively within the interfaces. The MID facilitates efficient access to comprehensive details related to the entity. More specifically, in Freebase, the MID is a unique identifier assigned to each entity, allowing for straightforward retrieval of entity-specific information. The friendly name of the MID is a natural language descriptor. For example, the MID of the friendly name \texttt{Jamaican} is $m.03\_r3$.
\subsection{Relation Filtering}
Through \textit{get\_relations}($e_i^{l}$), we obtain a candidate relation set $R$ associated with the initial entity in the question. Subsequently, DoG selects the optimal relation $\hat{r}_l$ from this set through in-context learning. The prompt and in-context examples are detailed in the \textit{In-context Learning} subsection of the appendix. Specifically, DoG first utilizes LLMs to identify the first-hop problem to be solved in the given question $q$. Then, it allows LLMs to choose the optimal relation according to the mentioned sub-question. This serves as a guiding principle for relation selection, avoiding the constant reliance on the complete multi-hop question throughout the entire reasoning stage, as seen in previous studies \cite{jiang2023structgpt, sun2024think}. We believe this short-sighted greedy strategy can guide a correct progression on the graph, alleviating the need to account for future inferential information regarding the multi-hop question. For example, for the question in Fig. \ref{fig:arc} ``In what year was the movie Joe Anderson starring in released'', the first-hop question to be addressed is ``Which film starred Joe Anderson?''. The linearized relation set is \{\texttt{$\sim$directed\_by}; \texttt{$\sim$starred\_actors}; \texttt{$\sim$written\_by}\} (“$\sim$” represents a passive relationship), from which the optimal relation \texttt{$\sim$starred\_actors} can be easily selected.

\subsection{Answer Trying}
After obtaining the optimal relation, our architecture invokes the triple-filling interface \textit{triple\_filling}($e_i^{l}$, $\hat{r}_l$) to acquire a complete triple, such as \texttt{\textless Joe Anderson, $\sim$starred\_actors, High Life\textgreater} in Fig. \ref{fig:arc}. Then, DoG utilizes LLMs to determine whether the retrieved triple can sufficiently support answering the question. If the triple is insufficient, DoG prompts LLMs to deeply contemplate the current question based on the provided triple. This allows DoG to generate answers based on a single triple, thus avoiding excessively long and potentially confusing paths composed of multiple triples. The prompt and in-context examples are detailed in the \textit{In-context Learning} subsection of the appendix. Notably, if the maximum iteration limit is reached without successfully generating an answer, the parameterized knowledge of LLMs is utilized to respond.

\subsection{Question Simplifying}
Once LLMs determine that a question is unanswerable with the current retrieved triple, it represents that further exploration is required. Inspired by how humans tackle complex questions, our architecture employs a question-simplifying strategy to transform questions from $K$ hop to $K$-1 hop based on the retrieved triple. Specifically, DoG utilizes a team of agents with distinct roles to engage in debate, ensuring the reliability of the reasoning process. The debate team consists of three roles.
\begin{table*}[htbp]
\centering
\begin{tabular}{c|c|c|ccc|c|c|c|c}
\toprule
\multirow{2}{*}{\textbf{Method}} & \multirow{2}{*}{\textbf{Class}} & \multirow{2}{*}{\textbf{LM}} & \multicolumn{3}{c|}{\textbf{MetaQA}}              & \multirow{2}{*}{\textbf{WebQSP}} & \multirow{2}{*}{\textbf{CWQ}} & \multirow{2}{*}{\textbf{WebQ}} & \multirow{2}{*}{\textbf{GrailQA}} \\ \cmidrule(lr){4-6}
            &   & & \textbf{1-hop} & \textbf{2-hop} & \textbf{3-hop} &    & &  &     \\ \midrule
KV-Mem      & \multirow{7}{*}{SL} & -     & 96.2    & 82.7    & 48.9    & 46.7      & 18.4   & -& -   \\
GraftNet    &   & -               & 97.0    & 94.8    & 77.7    & 66.4      & 36.8   & -& -   \\
PullNet     &   & -               & 97.0    & \underline{99.9}    & 91.4    & 68.1      & 45.9   & -& -   \\
EmbedKGQA   &   & RoBERTa         & \underline{97.5}    & 98.8    & 94.8    & 66.6      & -      & -& -   \\
NSM         &   & -               & 97.1    & \underline{99.9}    & 98.9    & 68.7      & 47.6   & -& -   \\
TransferNet &   & BERT            & 97.5    & \textbf{100.0} & \textbf{100.0} & \underline{71.4}      & \underline{48.6}   & -& -   \\
UniKGQA     &   & RoBERTa         & \textbf{98.0}  & \underline{99.9}    & \underline{99.9}    & \textbf{77.2}      & \textbf{51.2}   & -& -   \\ \midrule
StructGPT   & \multirow{4}{*}{ICL}     & GPT-3.5-Turbo   & 97.1    & 97.3    & 87.0    & 72.6      & -      & -& -   \\
KG-GPT      &   & GPT-3.5-Turbo   & 96.3    & 94.4    & 94.0    & -  & -      & -& -   \\
KB-BINDER   &   & Codex   & 93.5    & \textbf{99.9}  & \textbf{99.5}  & 74.4      & -      & -& 58.5\\ 
ToG         &   & GPT-3.5-Turbo   & -& -& -& 76.2      & \underline{57.1}   & 54.5    & 68.7\\ \midrule
DoG         & \multirow{4}{*}{ICL} & GPT-3.5-Turbo   & 98.6    & 96.6    & 90.9    & 88.6      & \textbf{58.2}   & \underline{78.2}    & \underline{77.8}\\
DoG         &   & Qwen-14B & 99.5    & 92.4    & 79.8    & 83.2      & 48.1   & 65.6    & 74.6\\
DoG         &   & Llama-3-8B      & \underline{99.8}    & 91.0    & 84.8    & \underline{90.2}      & 55.9   & 70.8    & 74.8\\
DoG         &   & GPT-4    & \textbf{100.0} & \underline{99.0}  & \underline{96.0}  & \textbf{91.0}      & 56.0   & \textbf{80.0}    & \textbf{80.0}    \\ \bottomrule            
\end{tabular}
\caption{Comparison with previous state-of-the-art Supervised Learning (SL) and In-Context Learning based methods. The best results for SL and ICL methods are marked in bold, and the second-best results are underlined. WebQ denotes the WebQuestions dataset. The ToG measurement on WebQSP is based on the F1 score rather than EM (Hits@1).} \label{tab:overall}
\end{table*}

\begin{itemize}
    \item Question simplifying expert (R1). This expert provides initial simplifications for questions, which may contain apparent errors. For example, the original question in Fig. \ref{fig:arc} is initially simplified as ``What are some notable films in which Joe Anderson has acted?". This is far from the intention of the original question.
    \item Critic (R2). The critic examines the simplification efforts of the above expert and offers suggestions for modifications. For instance, the above question is modified into ``When did High Life which was starred by Joe Anderson release?".
    \item Linguist (R3). This role ensures that the simplified question is not only semantically correct but also free from redundant information of previously resolved sub-questions. For example, the mentioned question is further refined to ``When did [High Life] release?".
\end{itemize}    

Due to the interdependency and progressive nature of the roles played by the three agents, DoG employs a one-by-one discussion strategy \cite{chan2023chateval}. Each agent, implemented by ChatGPT, takes turns contributing to the ongoing optimization of the simplified question, with the statements made by other agents serving as references for guiding subsequent remarks generation. After simplification, we obtain a slightly easier $K$-1 hop question, prompting LLMs to undergo iteration once again. In this way, the relation in the first-hop sub-question is removed in the simplified question, effectively avoiding the impact of false positive relations. The iteration process, from knowledge graph invocation to question simplification, continues until LLMs make an answerable decision in the answer-trying module. The prompt and in-context examples are shown in the \textit{In-context Learning} subsection of the appendix.

\section{Experiments}
\subsection{Dataset and Evaluation}
We select five public datasets to evaluate the reasoning ability over knowledge graphs: MetaQA \cite{zhang2018variational}, WebQSP \cite{yih2016value}, CWQ \cite{talmor2018web}, WebQuestions \cite{berant2013semantic}, and GrailQA \cite{gu2021beyond}. MetaQA comprises a movie ontology derived from the WikiMovies dataset \cite{miller2016key} and contains three sets of natural language question-answer pairs: 1-hop, 2-hop, and 3-hop. WebQSP contains questions sourced from the WebQuestions dataset, which are answerable using Freebase. CWQ is designed for answering complex questions that require reasoning over multiple web snippets. GraiQA, which tests three-level generalizations including i.i.d., compositional, and zero-shot, covers 3,720 relations and 86 domains from Freebase. Following \cite{xiong2024,sun2024think}, we uniformly sample 500 instances per type for the mentioned five datasets to reduce computational cost. We use \texttt{exact match accuracy} (Hits@1) to evaluate the reasoning performance of our framework and baselines following previous works \cite{jiang2023structgpt,xiong2024,sun2024think,baek2023knowledge}. For the experiment of integrating DoG with GPT-4, we uniformly sample only 100 instances per type from the mentioned datasets to reduce costs.

\subsection{Implementation Settings}
We preprocess the MetaQA dataset to construct a structured knowledge graph, facilitating subsequent query and retrieval operations. A local Virtuoso server is deployed for datasets derived from the Freebase. We utilize the OpenAI API to call ChatGPT (gpt-3.5-turbo-0125) and GPT-4 (gpt-4-0613). Additionally, we employ Qwen-14B and Llama-3-8B, running on 8 V100 GPUs, to verify the flexibility of DoG. The maximum number of debate rounds for the multi-agent team is limited to three, with only the best unique relation being recalled. We implement in-context learning across multiple modules: specifically, 10 exemplars for \textit{Relation Filtering} and \textit{Answer Trying}, and one exemplar for \textit{Question Simplifying}.
\begin{table*}[htbp]
\centering
\begin{tabular}{ccccccccc}
\toprule
\textbf{Num.} & \textbf{Settings} & \textbf{MetaQA}\textsuperscript{2} & \textbf{MetaQA}\textsuperscript{3} & \textbf{WebQSP} & \textbf{CWQ} & \textbf{WebQ} & \textbf{GrailQA} & \textbf{Avg.}\\ \midrule
1 & w/o SF and QS       & 76.6      & 38.8      & 77.4      & 43.0      & 67.8      & 69.3     & -     \\ 
2 & w/ SF and R1        & 91.4\ts{+14.8} & 83.4\ts{+44.6} & 81.0\ts{+3.6}  & 50.0\ts{+7.0}  & 69.8\ts{+2.0}  & 75.2\ts{+5.9} & +13.0 \\
3 & w/ SF, R1 and R2    & 90.6\ts{+14.0} & 85.2\ts{+46.6} & 83.6\ts{+6.2}  & 52.2\ts{+9.2}  & 72.2\ts{+4.4}  & \textbf{78.2}\ts{+8.9} & +14.9 \\
4 & w/ SF, R1, R2, and R3 & \textbf{96.6}\ts{+20.0} & \textbf{90.9}\ts{+52.1} & \textbf{88.6}\ts{+11.2} & \textbf{58.2}\ts{+15.2} & \textbf{78.2}\ts{+10.4} & 77.8\ts{+8.5} & \textbf{+19.6} \\
5 & w/ SF and QS'       & 86.6\ts{+10.0} & 68.2\ts{+30.6} & 81.4\ts{+4.0}  & 46.8\ts{+3.8}  & 71.2\ts{+3.4}  & 71.8\ts{+2.5} & +9.3  \\ \bottomrule
\end{tabular}
\caption{Ablation results. MetaQA\textsuperscript{\#} denotes the \#-hop split of this dataset. SF and QS refer to the subgraph focusing and question simplifying, respectively. R1, R2, and R3 are the different experts in QS. QS' indicates that the tasks of the mentioned three roles are fused into a single agent. Avg. represents the average performance increase across the datasets.} \label{tab:ablation}
\end{table*}

\subsection{Baselines}
Inspired by \cite{jiang2023structgpt}, we compare DoG with previous state-of-the-art supervised learning and in-context learning-based methods, to verify its effectiveness and superiority. Supervised learning: KV-Mem \cite{miller2016key}, GraftNet \cite{sun2018open}, PullNet \cite{sun2019pullnet}, EmbedKGQA \citep{saxena2020}, NSM \cite{he2021}, TransferNet \cite{shi2021}, UniKGQA \cite{JiangZ0W23}. In-context learning: StructGPT \cite{jiang2023structgpt}, KG-GPT \cite{kim2023kg}, KB-BINDER \cite{li2023few}, ToG \cite{sun2024think}. The baselines are detailed in the \textit{Baseline Introduction} subsection of the appendix.

\subsection{Reasoning on Knowledge Graphs}
\subsubsection{Main Result} Table \ref{tab:overall} presents a comparison across five public datasets. Taking GPT-3.5 as an example, we observe that DoG enables it to achieve competitive results on MetaQA and the best results on the other four datasets compared with baselines. Specifically, DoG outperforms the best-supervised method, UniKGQA, by 11.4\% on WebQSP. Additionally, it surpasses the best in-context learning method, ToG, by 23.7\% and 9.1\% on WebQuestions and GrailQA, respectively. These datasets comprise complex and compositional questions. Therefore, these results not only highlight the effectiveness and superiority of DoG but also confirm its capability for complex reasoning.

\subsubsection{Flexibility Verification} We conduct experiments on the aforementioned datasets to explore whether DoG enables other LLMs, including QWen, Llama, and GPT-4, to achieve complex reasoning on knowledge graphs. Experimental results in Table \ref{tab:overall} show that DoG facilitates improvements in some cases compared to GPT-3.5. Specifically, DoG with Llama achieves a 1.6\% improvement on WebQSP. It also allows GPT-4 to achieve the most significant improvement on the mentioned datasets. These results clearly demonstrate the flexibility and effectiveness of our architecture. We observe that the performance of DoG with Qwen is slightly lower than with other LLMs. This could be attributed to its marginally weaker complex reasoning capabilities compared to other LLMs.

\subsection{Ablation Studies}
\begin{figure}[!tbp]
\centering  
\includegraphics[width=0.8\linewidth]{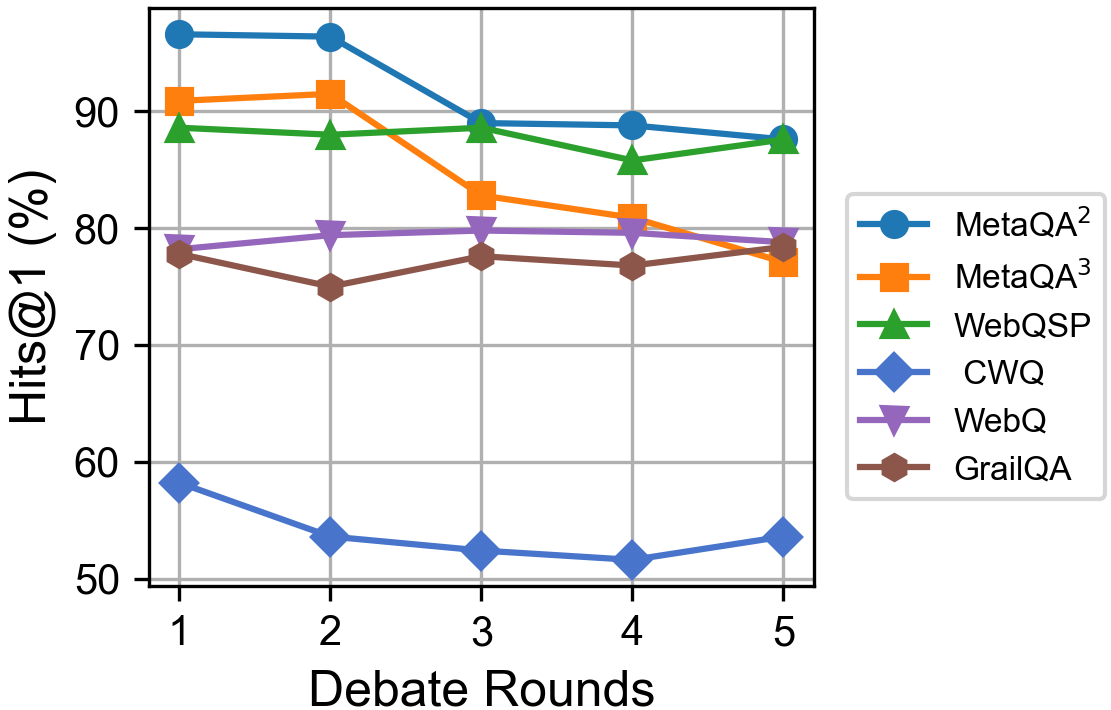}
\caption{Impact of debate rounds for LLM reasoning on knowledge graphs. It is unnecessary to simplify the question for the 1-hop question within MetaQA.}
\label{fig:debate_round}
\end{figure}
We conduct ablation experiments on the aforementioned datasets to analyze the contribution of each component of DoG. The ablation results for DoG with GPT-3.5 are presented in Table \ref{tab:ablation}. We perform experiments on the 2-hop and 3-hop splits of MetaQA, as the 1-hop questions do not require complex reasoning. Row 1 shows the results without the subgraph-focusing and question-simplifying components. In other words, this configuration allows LLMs to answer complex questions directly after collecting the whole set of evidence triples, rather than reasoning step by step. We observe a significant performance decrease compared to the results in Row 4, strongly demonstrating the effectiveness of the mentioned modules. Rows 2 and 3 aim to verify the contribution of the expert role in the debate team. The results show consistent improvements across five public datasets, suggesting that each agent plays a critical role in simplifying questions. This also highlights the importance of transforming complex questions into simpler ones for LLMs step-by-step reasoning on knowledge graphs. Row 5 aims to verify the necessity of the debating process, where the tasks of the three roles are performed by a single agent. The average result decreases by 10.3\% compared to Row 4, strongly supporting the effectiveness of the debating mechanism.
\begin{figure*}[!htbp]
\centering  
\includegraphics[width=0.82\linewidth]{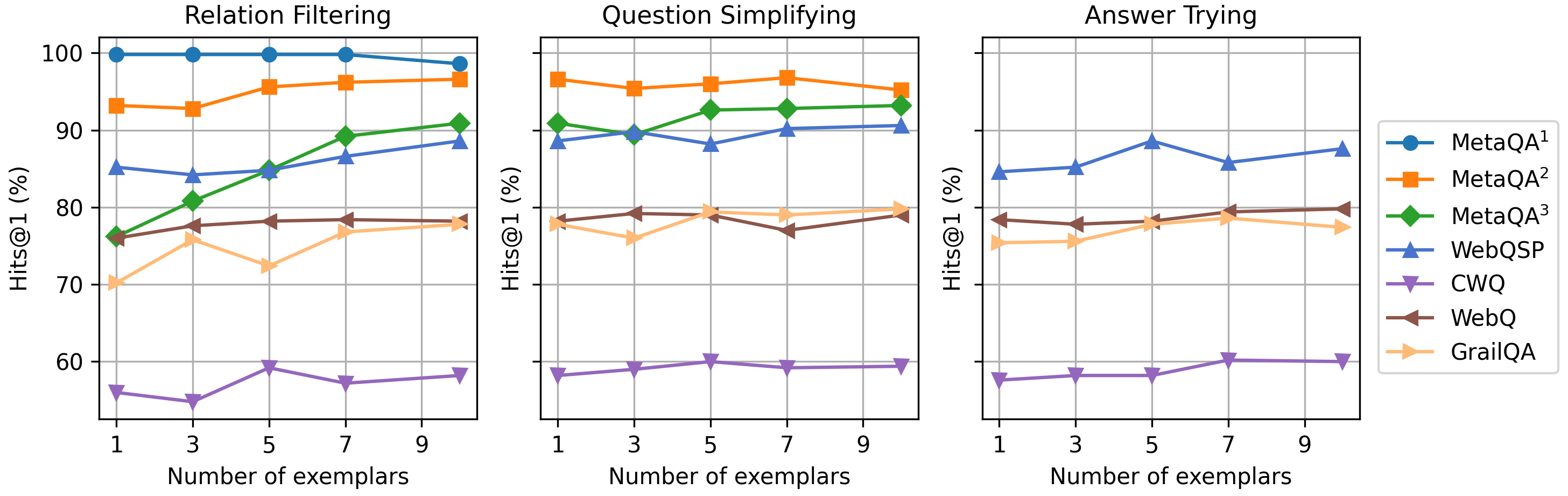}
\caption{Impacts of the number of exemplars on the performance of LLM reasoning. It is unnecessary to perform question simplifying for the 1-hop question within MetaQA. DoG does not utilize LLMs to generate answers for questions within MetaQA. Instead, it provides answers based on the last retrieved triple after iterative reasoning.}
\label{fig:examplar}
\end{figure*}
\begin{figure*}[!ht]
\centering  
\includegraphics[width=0.85\linewidth]{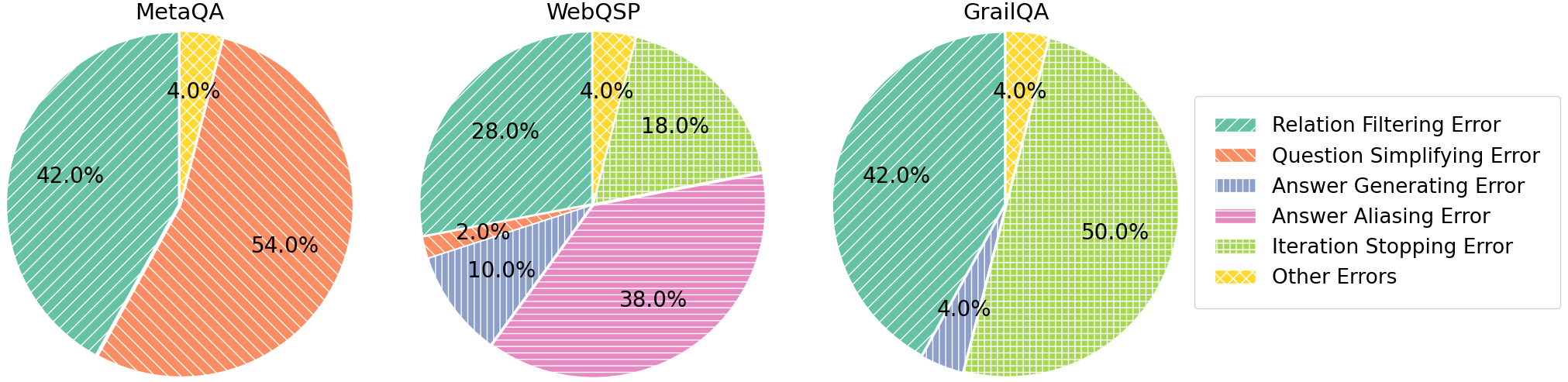}
\caption{Analysis of 50 sampled failure cases per dataset. We visualize the proportion of factors contributing to errors. We do not perform manual inspection for the failure cases in CWQ and WebQ due to the lack of annotations, such as those for the ground-truth relations.}
\label{fig:error}
\end{figure*}

\subsection{Analyses for Debate Rounds}
We conduct experiments to explore how the number of debate rounds affects LLM reasoning on knowledge graphs. Fig. \ref{fig:debate_round} shows the performance trend of DoG with GPT-3.5 as the number of debate rounds increases across the five datasets mentioned. We observe that DoG achieves the best results on the majority of datasets with just a single round of debates. Additionally, increasing the number of debate rounds leads to a performance decrease in some datasets. DoG utilizes a one-by-one discussion strategy, which makes each agent aware of the historical debate record. This makes the agents more susceptible to being influenced by the views of others, potentially leading to inaccurate decisions for question simplifications. We may also conclude that the agent is sufficiently strong to achieve the goal of instructions without needing iterative debates.

\subsection{Exemplar Impacts}
DoG leverages in-context learning to guide LLMs in performing relation filtering, question simplification, and answer trying during iterative reasoning. Specifically, DoG provides instructions and exemplars to help LLMs achieve these objectives. We conduct experiments on five public datasets to explore the impact of the number of exemplars on LLM reasoning. Fig. \ref{fig:examplar} shows the analyses for the mentioned three modules. In \textit{Relation Filtering}, we observe that reasoning performance improves as the number of exemplars increases in the majority of datasets. However, reasoning errors caused by relation filtering account for a large proportion, which we will discuss in the next subsection. In \textit{Question Simplifying}, the performance improvement is not significant with the increase in the number of exemplars, likely due to the complexity of this task. Converting questions from complex to simple step-by-step may be challenging for LLMs, and they may not be able to infer strategies for addressing this issue from exemplars. In \textit{Answer Trying}, we see that reasoning performance improves with the increase in the number of exemplars in most cases. In summary, the number of exemplars plays a critical role in decision-making, especially for less complex tasks. In contrast, for more complex tasks, detailed instructions may have a greater impact on LLM reasoning.

\subsection{Error Analyses}
To analyze the deficiency of DoG, we randomly select 50 failure cases from each dataset, including MetaQA, WebQSP, and GrailQA, for manual inspection. Fig. \ref{fig:error} shows the proportion of factors contributing to these errors. We observe that relation filtering errors are quite common. This may be caused by too many relations linked to the entities in questions, making it difficult for LLMs to accurately filter the most relevant relation. Iteration stopping errors denote LLMs make inaccurate decisions in the answer-trying module, either terminating the iterative reasoning too early or too late. This type of error is particularly prevalent in GrailQA cases. Answer aliasing errors mean the generated answers do not have the same description or wording as the annotations, even though they are semantically consistent. This error can be mitigated by introducing a rich collection of aliases. Answer generation errors refer to that LLMs provide incorrect answers based on accurately retrieved triples and simplified questions. Question simplifying errors represent that LLMs fail to transform questions from complex to easy. Additionally, other errors account for 4\% of the failure cases in each dataset. This type of error often occurs due to API access issues, an excessively long context, or exceeding the token limit per minute. More details can be found in the \textit{Failure Cases} subsection of the appendix.

\section{Conclusion and Future Work}
This paper proposes an iterative interactive framework, DoG, for knowledge graph question answering. It leverages the interactive learning and reasoning capabilities of LLMs to perform debating on knowledge graphs. Specifically, it employs a team of multi-role agents to transform questions from complex to simple, enabling LLMs to perform reliable step-by-step reasoning based on the retrieved knowledge triples. Extensive experiments across five public datasets demonstrate the effectiveness and superiority of DoG in the few-shot setting, outperforming nearly all in-context and supervised learning-based baselines. Additionally, the integration results with different LLMs verify its flexibility. In the future, we will explore enhancing relation filtering performance from knowledge graphs given the entity of questions.

\bibliography{aaai25}

\clearpage
\appendix
\section{Appendix}
\subsection{In-context Learning}
Table \ref{tab:prompt} shows the prompt, instruction, and exemplar utilized in the module within DoG.
\begin{table*}[htbp]
    \centering
    \begin{tabular}{p{0.9\textwidth}}
        \toprule
        \textbf{Prompt for Relation Filtering}\\
        For a multi-hop problem, solving it requires addressing several sub-problems that are logically preceded by the problem. Given a multi-hop problem and several relations, choose which relation should be used to solve the first subproblem. Here are some examples from which you should learn how to make your choices. It is worth noting that the ``$\sim$'' before a relationship means that the relationship is a passive relationship.\\
        \texttt{In-Context Few-shot}\\
        Question: Who acted in the movies written by the writer of Bottle Rocket?
        
        Relation\_set : relation\_1: \texttt{$\sim$ starred\_actors}, relation\_2: \texttt{written\_by}, relation\_3: \texttt{$\sim$ directed\_by}
        
        Explanation: This question contains three sub-questions
        
        1. First you need to find out  Bottle Rocket written by who.
        
        2. Then you need to find out what movies were written by that people.
        
        3. Finally you need to find out who acted in these movies.
        
        Therefore, the relationship between the sub-problems that need to be solved first should be relation\_2: written\_by,
        
        Output: relation\_2: \texttt{written\_by}\\
        \midrule
        \textbf{Prompt for Answer Trying}\\
        Given a question and the associated retrieved knowledge graph triples (entity, relation, entity), you are asked to answer whether it's sufficient for you to answer the question with these triples and your knowledge (Yes or No).\\
        \texttt{In-Context Few-shot}\\
        Question: Find the person who said ``Taste cannot be controlled by law'', what did this person die from?

        Knowledge Triples: ( Taste cannot be controlled by law., media\_common.quotation.author, Thomas Jefferson)
        
        Answer: \{No\}. Based on the given knowledge triples, it's not sufficient to answer the entire question. The triples only provide information about the person who said ``Taste cannot be controlled by law,'' which is Thomas Jefferson. To answer the second part of the question, it's necessary to have additional knowledge about where Thomas Jefferson's dead.
        
        Question: the species of qilin is the species of which fictional character?
        
        Knowledge Triples: ( Qilin, fictional\_universe.character\_species.characters\_of\_this\_species, Ki-rin)
        
        Answer: \{Yes\}. According to the knowledge triple provided, the species "Qilin" is associated with the fictional character species "Ki-rin" in fictional universes.\\
        \midrule
        \textbf{Prompt for Question Simplifying}\\
        \texttt{Role Description}\\
        \textbf{Question Simplifying Expert}
        \textit{You are an expert at problem simplification. What you are good at doing is answering the sub-problems of the original N-hop problem based on known information, so as to get new problems that need to be solved.}\\
        \textbf{Critic}
        \textit{You are a serious critic, please note: you need to compare [N-hop Question] and [Simplified\_question] obtained from the discussion in the chat log to see if they are the same. If they are the same, it means that the previous expert problem simplification failed and the task was not completed. You need to point this error out.}\\
        \textbf{Linguist}
        \textit{You are a linguist who is particularly good at dealing with irrelevant information in simplified problems. Regarding the simplification of collaborators’ output in chat records, if the simplification is not thorough, please provide a reasonable simplification solution. Specifically, it is necessary to ensure that there are no irrelevant constraints in the simplified question that originate from the answer entities of sub-questions that have already been answered in the original question.}\\
        \texttt{In-Context Few-shot}\\
        N-hop Question: Who directed films that share actors with the film Last Passenger?
        
        Knowledge triple: (Last Passenger, starred\_actors, Dougray Scott)
        
        Simplified\_question: Who directed the film that starred Dougray Scott?
        
        N-hop Question: Which films share the same director of Vampires Suck?
        
        Knowledge triple: (Vampires Suck, directed\_by, Jason Friedberg)
        
        Simplified\_question: Which film was directed by Jason Friedberg?\\
        \bottomrule
    \end{tabular}
    \caption{Prompt, instruction, and exemplar illustration utilized in the module within DoG.} \label{tab:prompt}
\end{table*}

\subsection{Baseline Introduction}
The brief introduction of supervised-based baselines is as follows.
\begin{itemize}
    \item KV-Mem \cite{miller2016key} is a key-value memory network that enhances the viability of reading knowledge sources, such as documents or knowledge bases, by utilizing different encodings during the addressing and output stages of the memory read operation. 
    \item GraftNet \cite{sun2018open} aims to provide answers based on a question-specific subgraph that includes text, entities, and relations. It employs heterogeneous update rules to handle knowledge base nodes differently from text nodes and utilizes a directed propagation method to constrain the propagation of embeddings within the graphs.
    \item PullNet \cite{sun2019pullnet} builds on the early GraftNet system but focuses on learning how to construct the subgraph. Unlike GraftNet, PullNet leverages a limited set of retrieval operations, with each operation expanding a graph node by acquiring new information from knowledge bases or corpora.
    \item EmbedKGQA \citep{saxena2020} is the first work that utilizes knowledge graph embeddings to perform multi-hop question answering over sparse knowledge graphs.
    \item NSM \cite{he2021} is a teacher-student network in which the student network aims to retrieve the correct answer to a question, while the teacher network learns intermediate supervision signals to enhance the reasoning capacity of the student network.
    \item TransferNet \cite{shi2021} is an effective and transparent model for multi-hop question answering that supports both label and text relations within a unified framework. TransferNet traverses entities across multiple steps. During each step, it focuses on different parts of the question, calculates activation scores for relations, and subsequently propagates the prior entity scores along these activated relations in a differentiable manner.
    \item UniKGQA \cite{JiangZ0W23} is a unified model designed for multi-hop question answering. It comprises a semantic matching module, leveraging a pre-trained language model for question-relation semantic matching, and a matching information propagation module that propagates this information along directed edges within knowledge graphs.
\end{itemize}

The brief description of in-context learning-based baselines is as follows.
\begin{itemize}
    \item StructGPT \cite{jiang2023structgpt} provides a specialized interface for gathering relevant evidence from structured data and utilizes large language models (LLMs) to focus on the reasoning task using the collected information. Specifically, it employs a linearization-generation procedure to enable LLMs to reason effectively on structured data, facilitated through the interface.
    \item KG-GPT \cite{kim2023kg} is a multi-purpose framework leveraging LLMs for tasks employing KGs. Specifically, it comprises three steps: sentence segmentation, graph retrieval, and inference, each aimed at partitioning sentences, retrieving relevant graph components, and deriving logical conclusions, respectively.
    \item KB-BINDER \cite{li2023few} is a unified, training-free framework that uses LLMs to generate logical forms for specific questions by mimicking a few demonstrations, and then grounds these forms on a knowledge base using BM25 score matching.
    \item ToG \cite{sun2024think} treats LLMs as agents that interactively explore relevant entities and relations on knowledge graphs, enabling them to perform reasoning based on the retrieved knowledge.
\end{itemize}

\subsection{Failure Cases}
\begin{figure}[tbp]
	\centering  
        \includegraphics[scale=0.4]{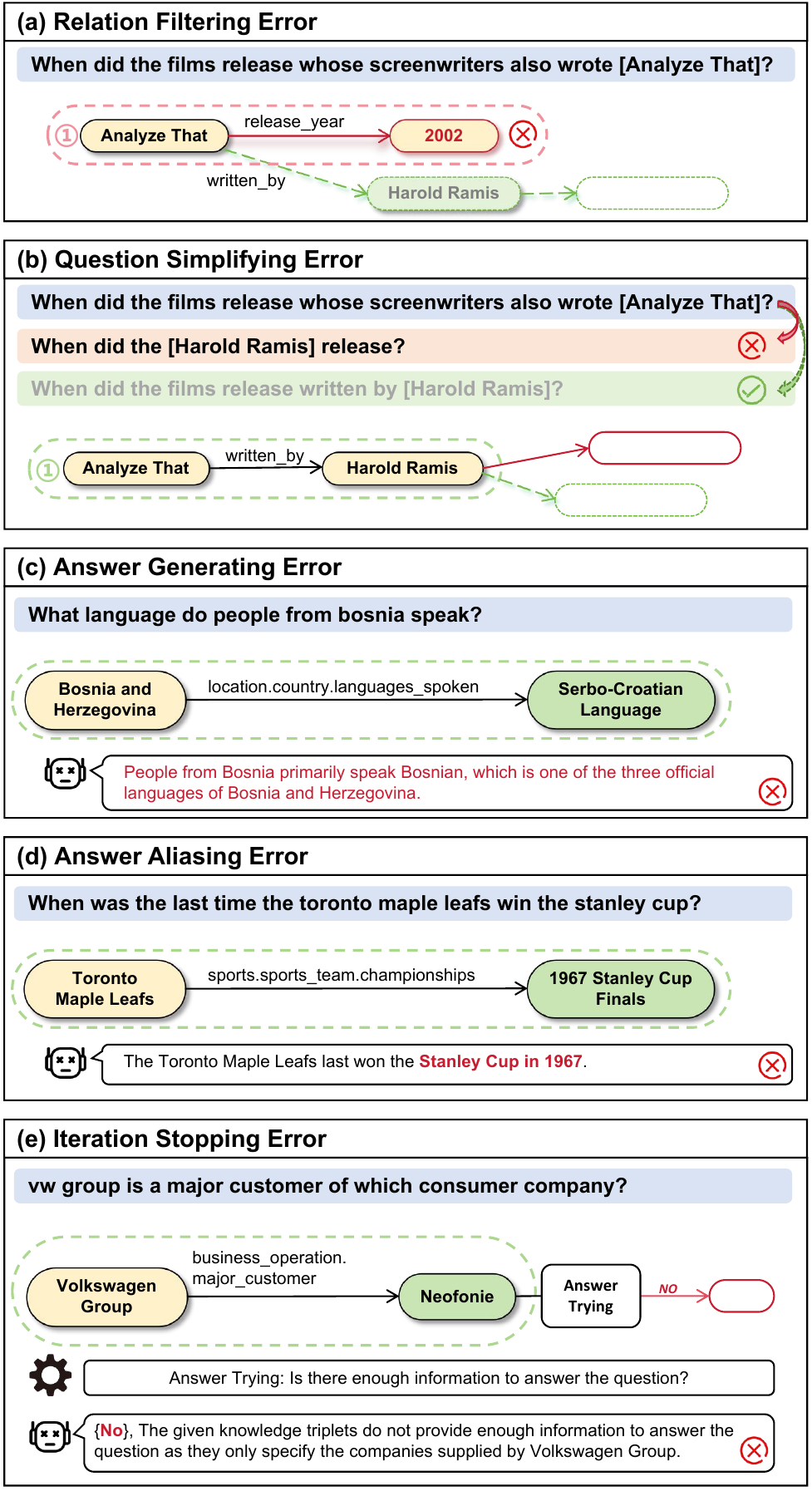}
	\caption{Failure cases.}
	\label{fig:error-case}
\end{figure}

Fig. \ref{fig:error-case} illustrates the error case that has been analyzed in the $\textit{Error Analysis}$ subsection.

\end{document}